\documentclass[letterpaper]{article} 
 \usepackage{aaai2027}
\usepackage[hyphens]{url}  
\usepackage{graphicx} 
\usepackage{natbib}  
\usepackage{caption} 
\usepackage{algorithm}
\usepackage{algpseudocode}

\usepackage{amsmath}
\usepackage{amssymb}
\usepackage{amsfonts}
\usepackage{bm} 

\usepackage{booktabs}
\usepackage{multirow}
\usepackage[table]{xcolor}
\usepackage{pifont}
\usepackage{enumitem}

\nocopyright 
\title{ARGen: Affect-Reinforced Generative Augmentation towards Vision-based Dynamic Emotion Perception}

\author{
    Huanzhen Wang\textsuperscript{\rm 1}, 
    Ziheng Zhou\textsuperscript{\rm 1}, 
    Jiaqi Song\textsuperscript{\rm 2}, 
    Li He\textsuperscript{\rm 1}, 
    Yunshi Lan\textsuperscript{\rm 2}, 
    Yan Wang\textsuperscript{\rm 2},
    Wenqiang Zhang\textsuperscript{\rm 1}
}
\affiliations{
    1. Fudan University, 2. East China Normal University
}

\begin{document}

\maketitle

\begin{abstract}
  Dynamic facial expression recognition in the wild remains challenging due to data scarcity and long-tail distributions, which hinder models from effectively learning the temporal dynamics of scarce emotions. To address these limitations, we propose \textbf{ARGen}, an \textbf{A}ffect-\textbf{R}einforced \textbf{Gen}erative Augmentation Framework that enables data-adaptive dynamic expression generation for robust emotion perception. ARGen operates in two stages: Affective Semantic Injection (ASI) and Adaptive Reinforcement Diffusion (ARD). The ASI stage establishes affective knowledge alignment through facial Action Units and employs a retrieval-augmented prompt generation strategy to synthesize consistent and fine-grained affective descriptions via large-scale visual-language models, thereby injecting interpretable emotional priors into the generation process. The ARD stage integrates text-conditioned image-to-video diffusion with reinforcement learning, introducing inter-frame conditional guidance and a multi-objective reward function to jointly optimize expression naturalness, facial integrity, and generative efficiency. Extensive experiments on both generation and recognition tasks verify that ARGen substantially enhances synthesis fidelity and improves recognition performance, establishing an interpretable and generalizable generative augmentation paradigm for vision-based affective computing.
\end{abstract}


\section{Introduction}

\begin{figure}[t]
  \centering
  \includegraphics[width=1.0\linewidth]{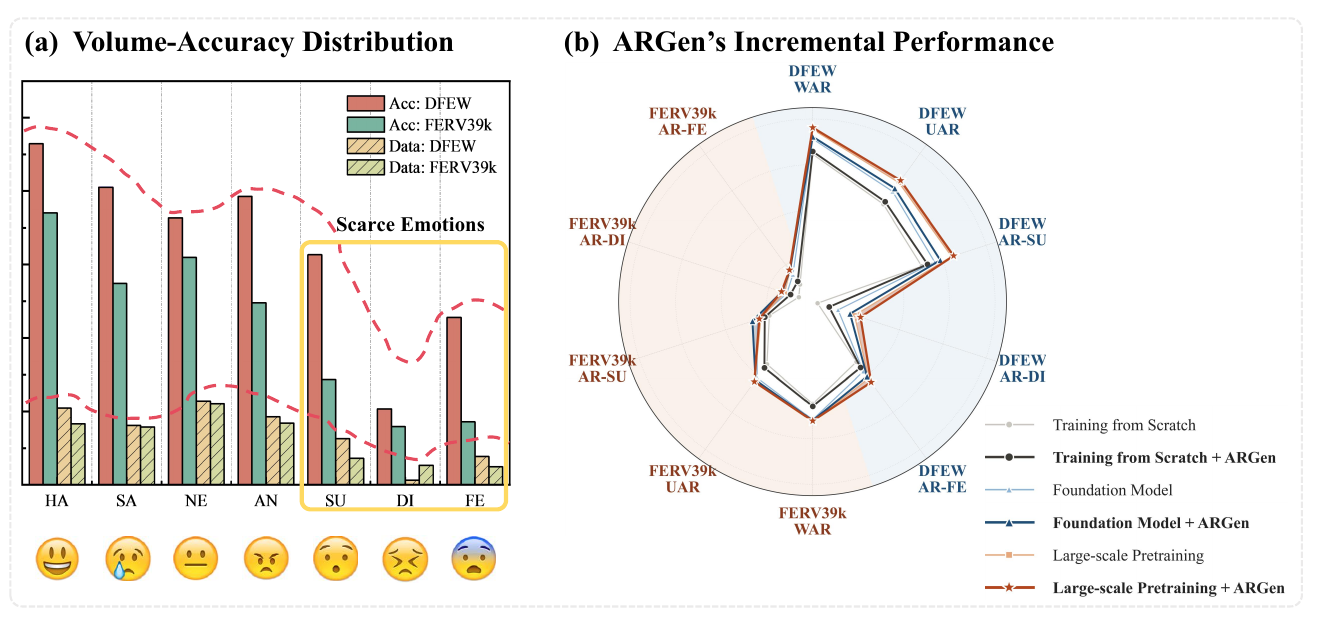}
  \caption{(a) demonstrates the long-tail distribution in terms of category sample counts and recognition accuracy in existing datasets. (b) demonstrates the incremental performance of our ARGen method in addressing the challenge of recognizing scarce facial expressions.}
  \label{motivation}
\end{figure}

Facial expressions convey rich emotional information vital for affective computing and human-computer interaction \cite{picard1997affective, calvo2010affect}. Compared with static images, dynamic expression videos capture temporal variations and emotional transitions, providing richer cues for complex emotion understanding \cite{wang2022systematic}. However, current dynamic facial expression recognition (DFER) methods \cite{zhao2021former,zhao2023prompting,cheng2024emotion} suffer from data limitations in real-world applications. Existing wild datasets are typically small-scale and highly imbalanced \cite{jiang2020dfew,wang2022ferv39k}, hindering models from learning discriminative temporal features for long-tail or scarce emotions, which degrades robustness and generalization (see Figure~\ref{motivation}).

To alleviate data scarcity, recent studies explore generative augmentation. However, most focus on static facial generation or editing \cite{pumarola2018ganimation, tu2023style}, neglecting temporal dynamics. Existing large-scale video generators and speaker-driven methods \cite{zhou2020makelttalk, guo2023animatediff} prioritize multimodal alignment over emotional semantic control. Furthermore, general diffusion models often introduce artifactual facial motions, including exaggerated amplitudes, spatial drift, and inter-frame discontinuity \cite{ho2022imagen, blattmann2023align, wang2023modelscope}, rendering them unsuitable for DFER enhancement. These challenges necessitate a dynamic expression generator that couples emotional priors with naturalness preservation and adaptive optimization.

Inspired by long-tail learning \cite{zhao2024ltgc,tian2022vl}, we propose ARGen (Affect-Reinforced Generative Augmentation), a dataset-adaptive dynamic expression generation framework requiring no external data. Specifically, the Affective Semantic Injection (ASI) mechanism constructs AU-based affective knowledge graphs for scarce categories \cite{ekman1978facial}. Utilizing a RAG-style retrieval module \cite{lewis2020retrieval}, ASI fuses semantic cues, reference frames, and emotion retrieval into a large Visual Language Model (VLM) \cite{bai2025qwen2} to generate interpretable prompts with coarse-to-fine consistency. Concurrently, the Adaptive Reinforcement Diffusion (ARD) mechanism formulates a Text-to-Image-to-Video (TI2V) pipeline via synergistic image-text collaboration, where inter-frame conditional guidance curves ensure coherent, natural expression transitions. Furthermore, reinforcement learning with multi-dimensional rewards (covering expression consistency, facial integrity, video quality, and generation steps) adaptively optimizes generation dynamics to balance naturalness, fidelity, and efficiency.

Unlike uniform enhancement or noise resampling across all categories \cite{tao2023freq,wang2025d2sp,wang2023rethinking}, ARGen applies targeted generative augmentation for scarce categories by explicitly incorporating affective semantic structures and reinforcement optimization into the generative process. ARGen synthesizes coherent, affectively consistent expression videos while interpretably fusing affective priors with generation strategies, providing robust data support for dynamic facial expression recognition.

In summary, our innovations and contributions are as follows:
\begin{itemize}[leftmargin=1em]
    \item We identify that the inherent long-tail nature of DFER datasets limits recognition performance on scarce emotion categories, and propose ARGen, the first adaptive generative augmentation framework for DFER requiring no external data.
    \item We introduce an AU-guided emotion prior injection method and an adaptive generative policy selection approach optimized via multi-scale rewards and reinforcement learning.
    \item Extensive experiments demonstrate that ARGen significantly outperforms multiple generation and recognition baselines, with comprehensive ablation studies validating the effectiveness of each component.
\end{itemize}

\section{Related Work}

\begin{figure*}[t]
  \centering
  \includegraphics[width=1\linewidth]{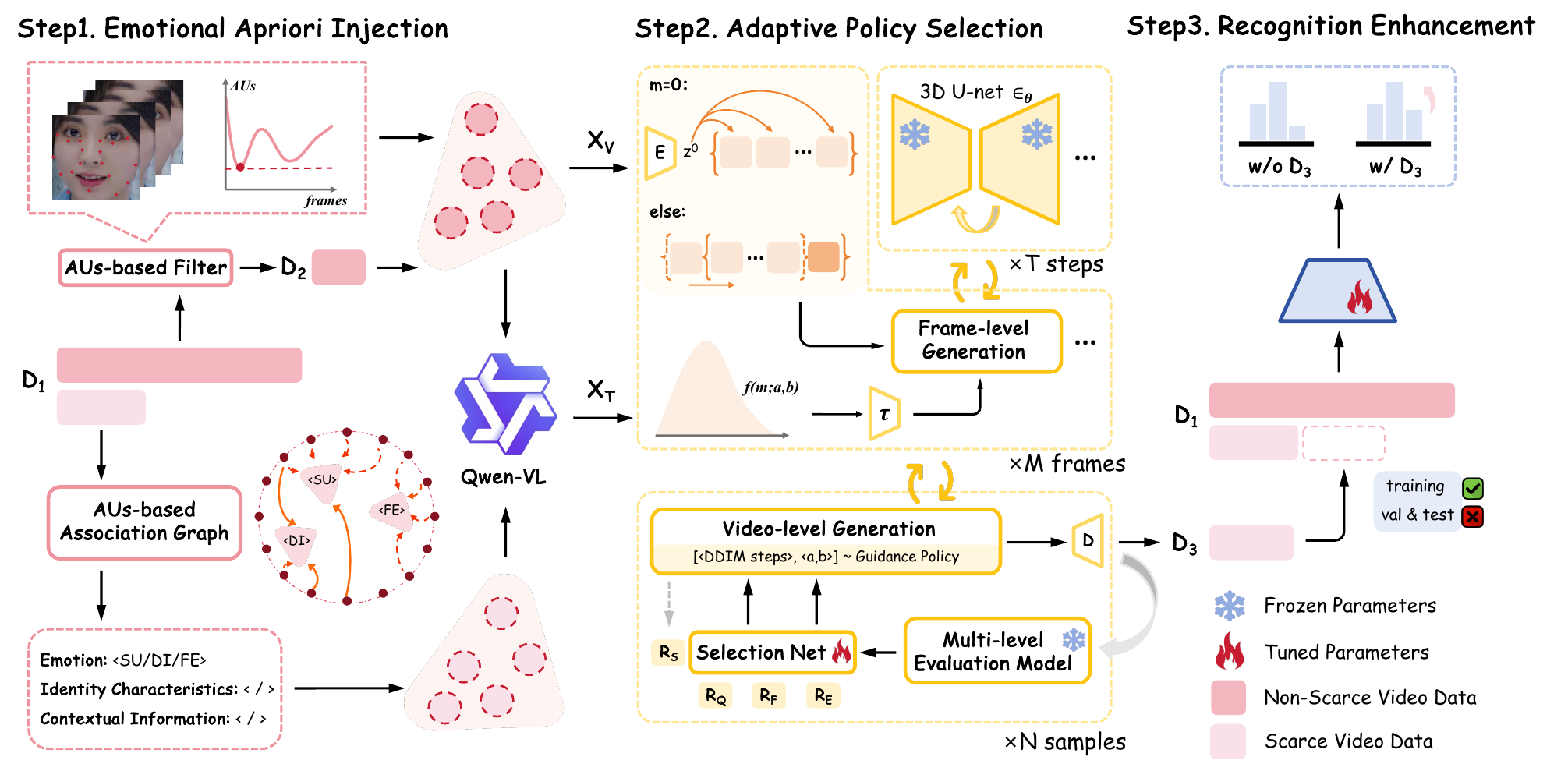}
  \caption{Overall framework of \textbf{ARGen}. The framework incorporates emotional priors derived from facial action units through vision-language models, followed by efficient video generation using diffusion models optimized via reinforcement learning.}
  \label{pipline}
\end{figure*}

\subsection{Dynamic expression recognition and generation}
Early facial expression recognition relied on handcrafted features in controlled environments \cite{livingstone2018ryerson}. Driven by deep learning and large-scale datasets \cite{li2020deep}, data-driven approaches now dominate, focusing on spatiotemporal dynamics to capture temporal variations. While in-the-wild datasets \cite{jiang2020dfew,wang2022ferv39k} establish DFER as a distinct task, existing methods \cite{zhao2021former,zhao2023prompting} struggle with underrepresented categories under long-tailed distributions. Unlike data noise processing \cite{wang2025d2sp} or sampling optimization \cite{tao2023freq}, our framework prioritizes targeted generative augmentation.

Furthermore, unlike talking-head generation \cite{zhang2024emodiffhead} or portrait editing \cite{guo2024liveportrait,ma2024follow} that emphasize identity or multimodal alignment, we focus on efficient dynamic expression generation with emotional priors. This directly addresses the interpretability and temporal scalability limitations of previous approaches \cite{bouzid2024facenhance,varanka2024towards}.

\subsection{Reinforcement learning in diffusion models}
Diffusion models (DM) \cite{ho2020denoising,dhariwal2021diffusion} have achieved state-of-the-art results across various tasks, including image generation \cite{esser2024scaling}, video generation \cite{bao2024vidu}, and image restoration \cite{xia2023diffir}, enabling flexible conditional generation across modalities like text and images. 

To align pre-training objectives with specific target intent, reinforcement learning (RL) \cite{sutton1998reinforcement} has been introduced to fine-tune diffusion models. Prior works \cite{wu2023human,wallace2023end,zhang2025adadiff} utilize human feedback or task-specific rewards to optimize text-image alignment, aesthetic appeal, and perceptual quality. In contrast, instead of uniform fine-tuning, our approach leverages RL to adaptively select efficient and controllable inference strategies tailored for dynamic emotion generation.

\subsection{Conditional image-to-video generation}
Conditional video generation synthesizes videos guided by user-provided signals, categorized by input modalities such as text-to-video (T2V) \cite{blattmann2023align,wang2023modelscope} and image-to-video (I2V) \cite{blattmann2021understanding,yang2018pose}. Since our work requires generating videos from dataset images guided by text prompts, we focus on text-conditioned image-to-video (TI2V) generation. While most existing TI2V approaches \cite{ni2024ti2v,preechakul2022diffusion} rely on general-purpose models, we mitigate the risk of overfitting on small datasets by prioritizing the injection of emotional priors through textual prompts.

\section{Method}
The main pipeline of ARGen comprises three stages (see Figure~\ref{pipline}). Based on empirical baselines, scarce expression categories—Surprise, Disgust, and Fear—are defined by a recognition accuracy below 50\% and a sample size under one-seventh of the total. Stage 1 injects emotional priors using visual language models and FACS. Stage 2 implements adaptive generation via diffusion models and reinforcement learning. Stage 3 applies recognition enhancement for downstream tasks. Crucially, ARGen integrates fine-grained emotional priors while optimizing the efficiency-fidelity tradeoff in dynamic generation. Below, we review diffusion preliminaries before detailing each stage's implementation.

\subsection{Preliminaries}
Diffusion models \cite{ho2020denoising,dhariwal2021diffusion} model data distribution via a Markov chain through forward noising and reverse denoising. Given $z_0 \sim q(z_0)$, the forward process sequentially adds Gaussian noise with a variance schedule $\{\beta_t\}_{t=1}^T$:
{\small
\begin{equation}
  q\left(z_{t} \mid z_{0}\right)=\mathcal{N}\left(\sqrt{\bar{\alpha}_{t}} z_{0},\left(1-\bar{\alpha}_{t}\right) I\right), \quad \bar{\alpha}_{t}=\prod_{i=1}^{t}\left(1-\beta_{i}\right)
\end{equation}
}
For a sufficiently large $T$, $z_T \approx \mathcal{N}(0, I)$. The reverse process parameterizes the true posterior as:
{\small
\begin{equation}
  p_{\theta}\left(z_{t-1} \mid z_{t}\right)=\mathcal{N}\left(\mu_{\theta}\left(z_{t}\right), \sigma_{t}^{2} I\right),
\end{equation}
}
where a network $\epsilon_\theta$ predicts the noise. Training minimizes the mean squared error:
{\small
\begin{equation}
\mathcal{L} = \mathbb{E}_{t, z_0, \epsilon} \left[ ||\epsilon - \epsilon_{\theta}(z_t, t)||^2_2 \right], \quad z_t \sim q(z_t | z_0)
\end{equation}
}
Conditional generation incorporates classifier-free guidance \cite{ho2022classifier} with a condition $y$. Randomly replacing $y$ with $\varnothing$ during training, sampling executes:
{\small
\begin{equation}
\hat{\epsilon}_{\theta}(z_{t}, t, y) = \epsilon_{\theta}(z_{t}, t, \varnothing) + g \cdot (\epsilon_{\theta}(z_{t}, t, y) - \epsilon_{\theta}(z_{t}, t, \varnothing)),
\label{eq1}
\end{equation}
}
where $g$ scales the guidance. Denoising progressively reconstructs $z_0$ from $z_T \sim \mathcal{N}(0,I)$.

ARGen builds upon a pre-trained image-to-video model \cite{wang2023modelscope,ni2024ti2v}, leveraging a ``repeat-slide'' queue and a ``replacement'' strategy. The initial frame latent $z_0 = E(x_0)$ is duplicated $K$ times into a queue $s_0 = \langle z_0, \dots, z_0 \rangle$. Generating one new frame at a time forces temporal attention to rely strictly on prior historical context. At each reverse step $t$, step-specific noise is added to $s_0$, and $s_t$ replaces the first $K$ frames of the current video latent $\hat z_t$ before denoising under condition $y$:
{\small
\begin{equation}
s_{t} \sim \mathcal{N}(\sqrt{\overline{\alpha}_{t}} s_{0}, (1 - \overline{\alpha}_{t})I), \quad \hat{z}_{t-1} \sim \mathcal{N}(\mu_{\theta}(\hat{z}_t, y), \sigma_t^2 I)
\end{equation}
}
The resulting clean frame latent $\hat z^K_0$ is decoded into pixels, enqueued, and the oldest frame is popped to form the next $s_0$, autoregressively completing the video.

To improve temporal stability, initial noise uses DDPM inversion instead of random noise: $s_0$ is perturbed through $T$ forward steps to yield $s_T$, which populates the first $K$ frames of the denoising latent body, while the last frame is initialized with $s_T^{K-1}$ for proximity approximation. A ``denoising-re-noising'' resampling strategy \cite{hoppe2022diffusion,lugmayr2022repaint} is integrated per step to refine motion coherence and fidelity.

\subsection{Emotional apriori injection}

\textbf{Dataset Decomposition and Identity Extraction.}
Given a dynamic sentiment dataset $D_1$, we partition it into a non-scarce subset $D_{ns}$ and a scarce subset $D_s$, denoting the neutral expressionless subset in $D_{ns}$ as $D^{*}_{ns}$. For each sequence $V_i \in D^{*}_{ns}$, we extract a representative identity image $I_i$ by minimizing the frame-level average AU intensity:
{\small
\begin{equation}
I_{i} = \arg\min_{m_{t} \in V_{i}} \frac{1}{K} \sum_{k=1}^{K} AU_{k}(m_{t}),
\end{equation}
}
where $AU_k(m_t)$ denotes the intensity of the $k$-th AU in frame $m_t$, and $K$ is the total number of AUs. The resulting identity image set is defined as $X_V = \{I_{1}, I_{2}, \cdots, I_{n}\}$.

\textbf{Knowledge Base Construction for Scarce Categories.}
To address data scarcity, we construct a specialized knowledge base $\mathcal{K} = \{(A_i, c_i, d_i)\}_{i=1}^{N}$, where each entry pairs an Action Unit (AU) set $A_i$ and its intensity scalars $r_{ij} \in [0,1]$ with an emotion label $c_i$ and a natural language description template $d_i$. Built upon Facial Action Coding System (FACS) literature \cite{ekman1978facial} and the scarce subset $D_s$, $\mathcal{K}$ focuses primarily on tail emotion categories and key AU combinations.

\textbf{AU Retrieval and Prompt Generation.}
Given an emotion label $e_t$ and an identity image $I$, we retrieve target emotion entries from the scarce knowledge base $\mathcal{K}$ and randomly sample $k$ candidate AU-description pairs. Each candidate is adjusted by the AU intensity vector $r$ to produce intensity-aware descriptions:
{\small
\begin{equation}
\{ (A^*_j, d^*_j) \}_{j=1}^k \sim \text{Uniform}(\{ (A_i, d_i) \mid c_i = e_t \}),
\end{equation}
\begin{equation}
F_{\text{AU}} = \{ \Phi_{\text{VL}}(d^*_j, r) = d^*_j + \text{modifier}(r) \}_{j=1}^k,
\end{equation}
}
where $\text{modifier}(r)$ injects intensity-related adverbs. Finally, the identity image, emotion label, and adjusted AU descriptions are fused via a vision-language model $\Phi_{\text{VL}}$ to generate the prompt. 

This process is repeated across all scarce categories for each reference image. We concatenate the descriptions $F_{\text{AU}}$ and strip explicit AU tags to construct the text prompt set $X_T$ (see Supplementary Materials Sec.C for implementation and code details).

\begin{algorithm}[tb]
\caption{Training and Adaptive Inference in ARD}
\label{alg:ARGen_ARD}
\textbf{Input}: Data pairs $\{(x_0, y)_i\}_{i=1}^N$, VLDM $\epsilon_\theta$, Encoders $\{E, \tau\}$, Action space $S = S_T \times S_{a \times b}$\\
\textbf{Output}: Optimized policy $\pi_w$, Augmented dataset $D_3$
\begin{algorithmic}[1]
\Statex \textbf{// Stage 1: Strategy Network Training}
\State $D_{train}, D_{val} \gets \text{Split}(\{(x_0, y)_i\}, 0.8)$
\Repeat
    \State $(x_0, y) \sim D_{train}$; $s = [E(x_0), \tau(y)]$
    \State $u = (T, a, b) \sim \pi_w(u|s)$ from $S_{cand} \subset S$
    \State \textbf{Dynamic Generation:} 
    \State $V \gets \text{Denoise}(z_T \sim \mathcal{N}(0, I), u)$ at step $t \in T$:
    \State \quad { $\hat{\epsilon}_{\theta}(z_{t_{s}}^{(m)})$ from eq.(\ref{eq11}) with cfg} \label{alg:eq_guidance}
    \State \textbf{Reward Auditing:} 
    \State $R(u) = R_S(u) + \lambda \sum_{k \in \{Q, F, E\}} R_k(u)$
    \State \quad If $R(u) < \text{top-}k \in D_{val}$, $R(u) \gets -\gamma$
    \State \textbf{Policy Update:} 
    \State $\nabla_w \mathcal{L} = \frac{1}{B} \sum_{j=1}^B R(u_j) \nabla_w \log \pi_w(u_j|s_j)$
\Until{Convergence}
\Statex
\Statex \textbf{// Stage 2: Adaptive Augmented Generation}
\State $D_3 \gets \emptyset$
\For{each $(x_0, y) \in \{(x_0, y)_i\}_{i=1}^N$}
    \State Compute $s = [E(x_0), \tau(y)]$
    \State Select $u^* = (T^*, a^*, b^*) = \arg\max_u \pi_w(u|s)$
    \State Generate $V^* = \text{Denoise}(z_{T^*}, u^*)$ using Line. \ref{alg:eq_guidance}
    \State $D_3 \gets D_3 \cup \{V^*\}$
\EndFor
\State \textbf{return} $D_3$
\end{algorithmic}
\end{algorithm}

\begin{table*}[t]
  \centering
  \renewcommand{\arraystretch}{1.4}
  \resizebox{1\linewidth}{!}{
  \begin{tabular}{cccccccccccc}
    \toprule
    \multirow{2}{*}{Method \& Architecture} & \multirow{2}{*}{Status}
    & \multicolumn{5}{c}{DFEW} & \multicolumn{5}{c}{FERV39k}\\
    \cmidrule(lr){3-7} \cmidrule(lr){8-12} 
&  & WAR & UAR & AR-SU & AR-DI & AR-FE & WAR & UAR & AR-SU & AR-DI & AR-FE \\
\midrule
\multicolumn{2}{c}{\textit{Training from Scratch}} & & & & & & & & &\\
\midrule
\multirow{2}{*}{Res18\_LSTM} & Original & 63.66 & 51.85 & 51.02 & 0.00 & 35.91 & 43.04 & 32.13 & 19.44 & 5.78 & 9.51 \\
  & \textbf{ARGen} & \textbf{65.22}(\textbf{1.56}$\uparrow$) & \textbf{53.25}(\textbf{1.40}$\uparrow$) & \textbf{52.80}(\textbf{1.78}$\uparrow$) & \textbf{6.90}\,(\textbf{6.90}$\uparrow$) & \textbf{38.58}\,(\textbf{2.67}$\uparrow$) & \textbf{44.46}\,(\textbf{1.42}$\uparrow$) & \textbf{34.42}\,(\textbf{2.29}$\uparrow$) & \textbf{21.67}\,(\textbf{2.23}$\uparrow$) & \textbf{11.85}\,(\textbf{6.07}$\uparrow$) & \textbf{11.23}\,(\textbf{1.72}$\uparrow$) \\
\midrule
\multirow{2}{*}{Res18\_Transformer} & Original & 66.59 & 55.60 & 52.82 & 5.52 & 36.92 & 45.96 & 35.29 & 22.10 & 11.78 & 11.83 \\
 & \textbf{ARGen} & \textbf{67.60}\,(\textbf{1.01}$\uparrow$) & \textbf{56.70}\,(\textbf{1.10}$\uparrow$) & \textbf{55.54}\,(\textbf{2.72}$\uparrow$) & \textbf{11.72}\,(\textbf{6.20}$\uparrow$) & \textbf{39.48}\,(\textbf{2.56}$\uparrow$) & \textbf{46.52}\,(\textbf{0.56}$\uparrow$) & \textbf{36.90}\,(\textbf{1.61}$\uparrow$) & \textbf{23.82}\,(\textbf{1.72}$\uparrow$) & \textbf{14.06}\,(\textbf{2.28}$\uparrow$) & \textbf{12.99}\,(\textbf{1.16}$\uparrow$) \\
\midrule
\multirow{2}{*}{VGG13\_LSTM} & Original & 62.47 & 49.44 & 47.62 & 0.00 & 22.10 & 43.95 & 32.44 & 17.08 & 0.64 & 4.18 \\
 & \textbf{ARGen} & \textbf{63.50}\,(\textbf{1.03}$\uparrow$) & \textbf{50.70}\,(\textbf{1.26}$\uparrow$) & \textbf{50.34}\,(\textbf{2.72}$\uparrow$) & \textbf{3.45}\,(\textbf{3.45}$\uparrow$) & \textbf{25.97}\,(\textbf{3.87}$\uparrow$) & \textbf{45.07}\,(\textbf{1.12}$\uparrow$) & \textbf{34.01}\,(\textbf{1.57}$\uparrow$) & \textbf{18.50}\,(\textbf{1.42}$\uparrow$) & \textbf{4.93}\,(\textbf{4.29}$\uparrow$) & \textbf{6.26}\,(\textbf{2.08}$\uparrow$) \\
\midrule
\multirow{2}{*}{VGG13\_Transformer} & Original & 65.50 & 53.94 & 49.66 & 3.45 & 35.67 & 46.09 & 35.88 & 20.85 & 7.07 & 11.60 \\
 & \textbf{ARGen} & \textbf{66.77}\,(\textbf{1.27}$\uparrow$) & \textbf{55.53}\,(\textbf{1.59}$\uparrow$) & \textbf{53.22}\,(\textbf{3.56}$\uparrow$) & \textbf{8.28}\,(\textbf{4.83}$\uparrow$) & \textbf{38.89}\,(\textbf{3.22}$\uparrow$) & \textbf{47.17}\,(\textbf{1.08}$\uparrow$) & \textbf{38.11}\,(\textbf{2.23}$\uparrow$) & \textbf{23.98}\,(\textbf{3.13}$\uparrow$) & \textbf{9.59}\,(\textbf{2.52}$\uparrow$) & \textbf{13.58}\,(\textbf{1.98}$\uparrow$) \\
\midrule
\multicolumn{2}{c}{\textit{Foundation Model-based Adaptation}} & & & & & & & & &\\
\midrule
\multirow{2}{*}{DFER-CLIP \cite{zhao2023prompting}} & Original & 71.25 & 59.61 & 56.25 & 11.72 & 37.81 & 51.65 & 41.27 & 25.39 & 11.78 & 14.62 \\
 & \textbf{ARGen} & \textbf{72.22}\,(\textbf{0.97}$\uparrow$) & \textbf{61.26}\,(\textbf{1.65}$\uparrow$) & \textbf{58.84}\,(\textbf{2.59}$\uparrow$) & \textbf{17.24}\,(\textbf{5.52}$\uparrow$) & \textbf{40.54}\,(\textbf{2.73}$\uparrow$) & \textbf{52.04}\,(\textbf{0.39}$\uparrow$) & \textbf{42.47}\,(\textbf{1.20}$\uparrow$) & \textbf{27.68}\,(\textbf{2.29}$\uparrow$) & \textbf{14.05}\,(\textbf{2.27}$\uparrow$) & \textbf{16.77}\,(\textbf{2.15}$\uparrow$) \\
\midrule
\multicolumn{2}{c}{\textit{Large-scale Pre-training with Additional Data}} & & & & & & & & &\\
\midrule
\multirow{2}{*}{S4D \cite{chen2025static}} & Original & 75.45 & 64.37 & 64.04 & 19.31 & 41.64 & 52.41 & 42.96 & 23.81 & 12.83 & 16.71 \\
 & \textbf{ARGen} & \textbf{76.23}\,(\textbf{0.78}$\uparrow$) & \textbf{65.57}\,(\textbf{1.20}$\uparrow$) & \textbf{65.01}\,(\textbf{0.97}$\uparrow$) & \textbf{22.07}\,(\textbf{2.76}$\uparrow$) & \textbf{43.60}\,(\textbf{1.96}$\uparrow$) & \textbf{52.19}\,(\textbf{0.22}$\downarrow$) & \textbf{43.32}\,(\textbf{0.36}$\uparrow$) & \textbf{24.52}\,(\textbf{0.71}$\uparrow$) & \textbf{14.23}\,(\textbf{1.40}$\uparrow$) & \textbf{17.12}\,(\textbf{0.41}$\uparrow$) \\
\midrule
\rowcolor{gray!10}
\multicolumn{2}{c}{\textbf{Average Improvement}} & \textbf{1.10} & \textbf{1.37} & \textbf{2.39} & \textbf{4.94} & \textbf{2.84} & \textbf{0.73} & \textbf{1.54} & \textbf{1.92} & \textbf{3.14} & \textbf{1.58} \\
    \bottomrule
  \end{tabular}
  }
  \caption{ARGen's performance comparison (\%) of different recognition baselines on DFEW and FERV39k.}
  \label{rec1}
\end{table*}

\subsection{Adaptive policy selection}
The comprehensive procedure is detailed in Algorithm~\ref{alg:ARGen_ARD}.

\textbf{State space and action space.}
Given a reference image $x_0 \in X_V$ and text prompt $y \in X_T$, a conditional video diffusion model synthesizes $M$ frames $V = \{V_m\}_{m=1}^M$. We introduce a parameter selection policy network atop the latent diffusion model (LDM) to jointly determine three inference hyperparameters: the inversion steps $T \in S_T$ and the text guidance function parameters $(a,b) \in S_{a \times b}$. Defining the text encoder as $\tau$ and the image encoder-decoder as $(E, D)$, the state representations are encoded as $c = \tau(y)$ and $r = E(x_0)$. The policy network maps $[c, r]$ to three action sets. Executing the selected action $u = (T, a, b)$, DDIM inversion operates with frame-wise guidance weights $f(m;a, b)$ over $T$ denoising steps to generate the video:
{\small
\begin{equation}
V = D(\mathrm{LDM}_{\mathrm{DDIM}}^{(y, x_0) \rightarrow (T, a, b)}(Z_{\hat{t}}, c, r; f(m; a, b))),
\end{equation}
}
where $\hat t=\{t_1,\dots,t_T\}$ denotes the sampling time-step sequence, and $Z_{\hat t}=\{z_{t_T}^{(m)}\}_{m=1}^M$ represents the initialized latent stack with $z_{t_T}^{(m)}\sim\mathcal{N}(0,I)$. Sampling follows classifier-free guidance (CFG), interpolating unconditional and conditional noise predictions.

{\small
\begin{equation}
\hat{\epsilon}_{\theta}(z_{t_{s}}^{(m)}) = \epsilon_{\theta}(z_{t_{s}}^{(m)}; \varnothing) + f(m) [\epsilon_{\theta}(z_{t_{s}}^{(m)};c,r)-\epsilon_{\theta}(z_{t_{s}}^{(m)}; \varnothing)]
\label{eq11}
\end{equation}
}

To dynamically modulate emotional intensity, we implement a Beta-like function $f(m;a,b)$ to control the diffusion guidance factor across frames. This design ensures smooth, physiologically consistent control, facilitating natural transitions from neutral to peak expressions while preserving sampling stability and temporal coherence.

{\small
\begin{equation}
f(m;a,b) =g_{min}+(g_{max}-g_{min})\cdot {(\frac{m}{M})^{a-1} (1-\frac{m}{M})^{b-1}}, 
\end{equation}
}

where $g_{min}$ and $g_{max}$ represent the minimum and maximum values of the guidance strength respectively.

\textbf{Policy.}
The policy network processes $[c,r]$ to output three-headed logits $s_T, s_a, s_b$, defining probability distributions over their respective discrete action spaces. The joint policy is factorized as:
\begin{equation}
\pi(u \mid x_0, y) = \pi_T(T \mid c, r) \cdot \pi_a(a \mid c, r) \cdot \pi_b(b \mid c, r).
\end{equation}
Architecturally, the policy network employs a self-attention mechanism to fuse condition $c$ and reference $r$, followed by independent multi-layer perceptron (MLP) heads. Actions are sampled from $\pi$ during training, while deterministic $\arg\max$ decoding is applied during inference.

\textbf{Reward.}
For a generated video $V = \{V_m\}_{m=1}^M$, $R_S(u)$ denotes the normalized remaining step size relative to the maximum steps in $S_T$. $R_Q(u)$ balances intra-frame quality (sharpness and contrast) and inter-frame consistency (smoothness). Following \cite{tu2021ugc}, we set the weighting parameter $\alpha = 0.6$, where $Q_m$ and $Q_M$ represent per-frame and mean video quality metrics. $R_F(u)$ calculates the proportion of frames containing faces based on the face presence probability $p_m$. To penalize exaggerated facial motions, the expression amplitude reward $R_E(u)$ measures intensity via the $L_2$ norm of the per-frame AU vector, normalized against an empirical maximum amplitude $a_{max}$. Implementation details for these rewards are provided in Supplementary Materials Sec.D.

{\small
\begin{equation}
R_S(u) = 1 - \frac{s_T}{S_{Tmax}}
\end{equation}
\begin{equation}
R_Q(u) = \alpha Q_m+(1-\alpha)Q_M
\end{equation}
\begin{equation}
R_F(u) = \frac{1}{M} \sum_{m=1}^{M} p_m
\end{equation}
\begin{equation}
R_E(u) = \frac{1}{M} \sum_{m=1}^{M} \max\Big(0,\, 1 - \frac{\|{a}_m\|_2}{a_{\max}}\Big)
\end{equation}
}

To ensure fairness given prompt $y$ and reference $x_0$, a small Cartesian candidate pool is sampled across three sets to generate videos for scoring. During training, candidates ranking above the top-$k$ threshold receive rewards, while others incur a penalty $\gamma$. Concurrently, the final reward is computed as a weighted sum of these components.

{\small
\begin{equation}
R(u) = \begin{cases} R_S(u) + \lambda (R_Q(u) + R_F(u)+R_E(u)), & normal, \\ -\gamma, & else. \end{cases}
\end{equation}
}

\begin{table}[t]
\centering
\renewcommand{\arraystretch}{1.2}
\resizebox{1\linewidth}{!}{
\begin{tabular}{cccc}
\toprule
Method & FVD$\downarrow$ & sFVD$\downarrow$ & tFVD$\downarrow$\\
\midrule
DynamiCrafter \cite{xing2024dynamicrafter} & 170.16 & 345.16 $\pm$ 137.32 & 208.35 $\pm$ 45.79 \\
TI2V-Zero \cite{ni2024ti2v} & 81.72 & 170.20 $\pm$ 57.90 & 109.41 $\pm$ 47.26\\
\midrule
\rowcolor{gray!10}
\textbf{ARGen(ours)} & \textbf{56.75} & \textbf{133.62 $\pm$ 50.94} & \textbf{81.21 $\pm$ 27.07}\\
\bottomrule
\end{tabular}
}
\caption{ARGen's quantitative comparison of different methods for generation on CK+.}
\label{gen1}
\end{table}

\textbf{Target optimization.}
The optimization objective is to maximize the expected reward:
{\small
\begin{equation}
\max_{w} \mathcal{L} = \mathbb{E}_{u \sim \pi(\cdot | y, x_0)}[R(u)]
\end{equation}
}
In this paper, we use the policy gradient method \cite{sutton1998reinforcement} to learn the parameters $w$ for the selection network. The expected gradient can be derived as follows:
{\small
\begin{equation}
\nabla_{w} \mathcal{L} = \mathbb{E} \left[ R(u) \nabla_{w} \log \pi(u | y, x_{0}) \right],
\end{equation}
}
which approximates in small batches and utilizes a factorization strategy to decompose the log-likelihood into a sum of three terms:
{\small
\begin{equation}
\nabla_{w} \mathcal{L} \approx \frac{1}{B} \sum_{j=1}^{B} R(u_{j}) \nabla_{w} ( \sum_{i}^{T,a,b}\log \pi_{i}(i_{j} \mid c_{j}, r_{j})),
\end{equation}
}

where $B$ denotes the mini-batch size of condition-reference pairs. We apply the Adam optimizer to update the network, balancing inference speed, generation quality, and facial fidelity. During inference, the optimal generation steps and guidance parameters are adaptively determined via $\arg\max$ decoding: $u^*=(T^\star, a^\star, b^\star) = \arg\max_{u} \pi(u|c,r)$, enabling dynamic per-sample inference.

\begin{figure*}[t]
  \centering
  \includegraphics[width=1\linewidth]{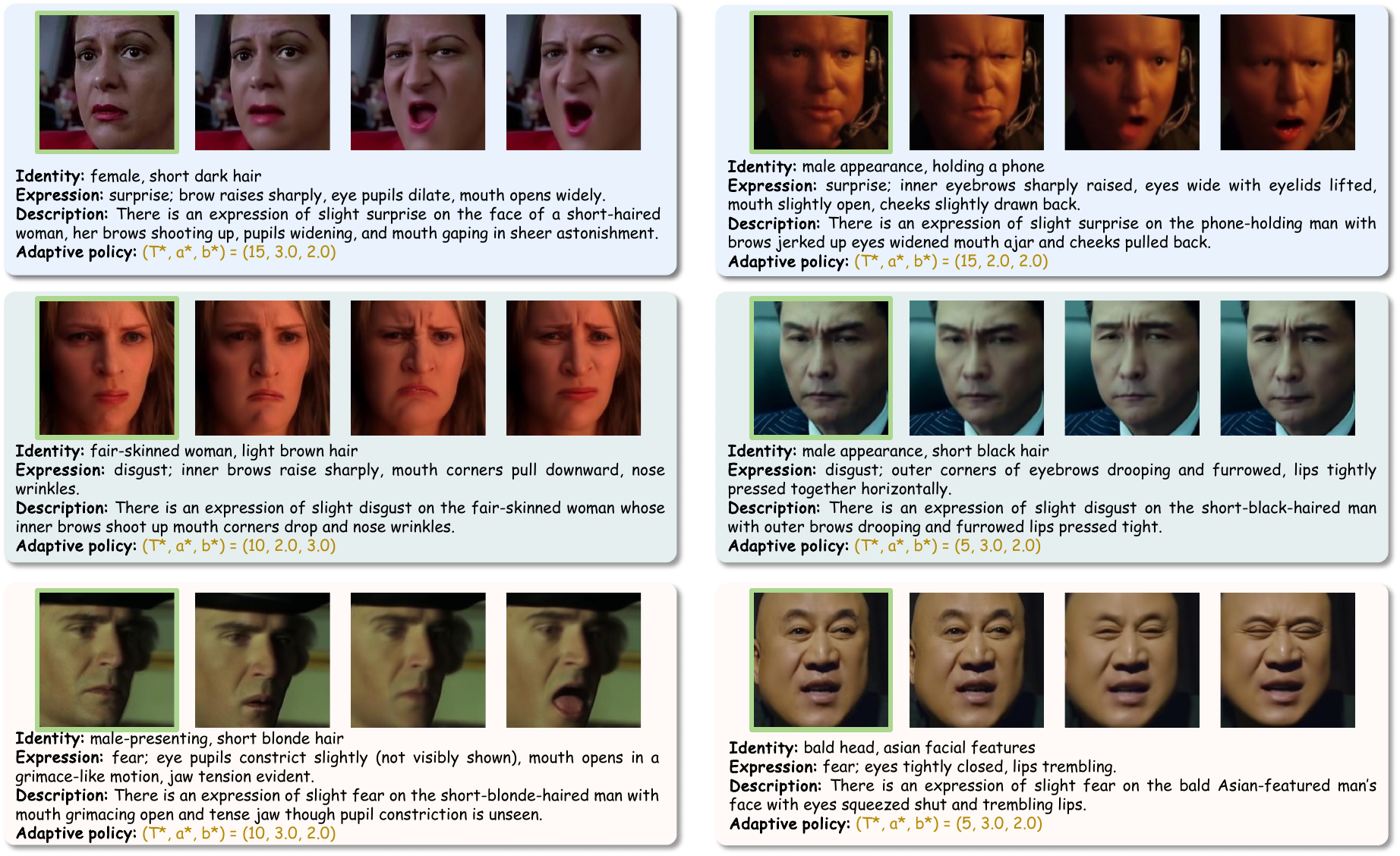}
  \caption{Visualization of generation performance for scarce categories in DFEW and FERV39k. The green box indicates the input neutral reference frame and brown markers represent adaptive policy selection results for the input in ARD. We generated 16 video frames throughout the experiments and the four frames shown correspond to frames 1, 6, 11, and 16. The three sections, from top to bottom, represent the emotions of surprise, disgust, and fear.}
  \label{vis0}
\end{figure*}

\section{Experiment}

\subsection{Datasets and metrics}
\textbf{Datasets.} We conducted comprehensive evaluation across three benchmarks in generation and recognition stages (see Supplementary Materials Sec.B for dataset details).
The zero-shot metrics for the generation phase are primarily reflected in CK+ \cite{lucey2010extended} dataset. The metrics for the recognition phase are primarily reflected in DFEW \cite{jiang2020dfew} and FERV39k \cite{wang2022ferv39k}.

\textbf{CK+} is a laboratory-controlled benchmark containing 593 sequences across 123 subjects, with 327 sequences annotated with seven basic emotions transitioning from neutral to peak expressions.

\textbf{DFEW} is a large-scale, in-the-wild dataset comprising over 11K movie and TV video clips, featuring diverse poses, illuminations, and occlusions to evaluate real-world robustness.

\textbf{FERV39k}  consists of approximately 39K unconstrained video samples from interviews and social media spanning seven emotions. With high scene variability and class imbalance, it benchmarks long-tail recognition and affective video generation.

\textbf{Data Preprocessing.} For generation, inputs are resized to 256$\times$256 with 16 uniformly sampled frames to form fixed-length ground-truth clips. For recognition, frames are center-cropped to 224$\times$224 for evaluation, ensuring augmentations strictly modify the training set. DFEW and FERV39k adopt 5-fold cross-validation and the default split, respectively, while CK+ serves as a zero-shot validation benchmark. The policy network partition follows an 8:2 train-test split.

\textbf{Metrics.} Following \cite{ho2022imagen,ni2024ti2v,zhao2023prompting,wang2025d2sp}, we evaluate video quality, temporal consistency, and diversity via Fr\'echet Video Distance (FVD), sFVD, and tFVD. Specifically, sFVD computes the distance between real and synthetic videos of the same subject, whereas tFVD measures alignment under identical text inputs for the same emotion. For downstream recognition, we adopt Unweighted Average Recall (UAR) and Weighted Average Recall (WAR).

\subsection{Implementation details}
\textbf{Model and Hardware Configuration.} We utilize Qwen2.5-7B-VL \cite{bai2025qwen2} as the vision-language prior model in ASI and modify ModelScopeT2V \cite{wang2023modelscope} as our foundational pipeline in ARD. Experiments are accelerated using two NVIDIA 3090 and two NVIDIA A6000 GPUs.

\begin{table}[t]
\centering
\renewcommand{\arraystretch}{1.15}
\resizebox{1\linewidth}{!}{
\begin{tabular}{cccccc}
\toprule
Ablation & WAR & UAR & AR-SU & AR-DI & AR-FE\\
\midrule
Original(w/o TA)& 64.50 & 52.83 & 50.34 & 0.00 & 34.96 \\
Original(w/ TA)& 66.59 & 55.60 & 52.82 & 5.52 & 36.92\\
\midrule
ARGen(w/o ASI) & 67.31 & 56.39 & 55.10 & 6.90 & 35.36 \\
$\Delta$ & +1.08\% & +1.42\% & +4.13\% & +25.00\% & -4.23\% \\
\midrule
ARGen(w/o ARD) & 67.06 & 54.98 & 54.79 & 6.21 & 37.73 \\
$\Delta$ & +0.71\% & -1.12\% & +3.73\% & +12.5\% & +2.19\% \\
\midrule
\rowcolor{gray!10}
\textbf{ARGen} & \textbf{67.60} & \textbf{56.70} & \textbf{55.54} & \textbf{11.72} & \textbf{39.48}\\
\rowcolor{gray!10}
$\Delta$ & \textbf{+1.52\%} & \textbf{+1.98\%} & \textbf{+5.15\%} & \textbf{+112.32\%} & \textbf{+6.93\%}\\
\bottomrule
\end{tabular}
}
\caption{Ablation study on two-stage performance comparison (\%) of ARGen on DFEW  using Res18\_Transformer. “TA” refers to the traditional image data augmentation methods we use in our setup.}
\label{rec2}
\end{table}

\textbf{Hyperparameter.} In ASI, Qwen2.5-7B-VL generates English prompts with a maximum length of 256, temperature of 0.8, and $\text{top\_p} = 0.9$. In ARD, the policy action spaces are defined as $S_T \in \{5,10,15,20\}$ and $S_{a \times b} \in \{(2.0, 2.0), (3.0, 2.0), (2.0, 3.0)\}$, with reward penalty $\gamma = 1$. Consistent with \cite{ni2024ti2v}, guidance parameters bound at $g_{\min}=7$ and $g_{\max}=11$. During RL optimization, the action space is randomly sampled four times per iteration with a learning rate of $10^{-4}$, batch size of 32, and quality weight $\lambda=3$. For downstream evaluation, all recognition baselines are optimized via SGD with a fixed learning rate of $10^{-3}$. Both LSTM and Transformer baselines employ two layers with a hidden dimension of 256.

\begin{table}[t]
\centering
\renewcommand{\arraystretch}{1.15}
\resizebox{0.95\linewidth}{!}{
\begin{tabular}{ccccc}
\toprule
\multicolumn{2}{c}{Rewards} & \multirow{2}{*}{FVD$\downarrow$} & \multirow{2}{*}{sFVD$\downarrow$} & \multirow{2}{*}{tFVD$\downarrow$} \\
\cmidrule(lr){1-2}
$R_Q$ & $R_F,R_E$ &  &  & \\
\midrule
\ding{55} & \ding{55} & 85.87 & 187.55 $\pm$ 75.36 & 112.98 $\pm$ 48.22 \\
\ding{51} & \ding{55} & 65.39 & 164.19 $\pm$ 62.72 & 102.28 $\pm$ 46.15\\
\ding{55} & \ding{51} & 78.96 & 203.93 $\pm$ 78.24 & 119.44 $\pm$ 60.51 \\
\rowcolor{gray!10}
\midrule
\ding{51} & \ding{51} & \textbf{56.75} & \textbf{133.62 $\pm$ 50.94} & \textbf{81.21 $\pm$ 27.07}\\
\bottomrule
\end{tabular}}
\caption{ARGen's quantitative comparison of different rewards for generation on CK+.}
\label{gen2}
\end{table}

\begin{figure}[t]
  \centering
  \includegraphics[width=1\linewidth]{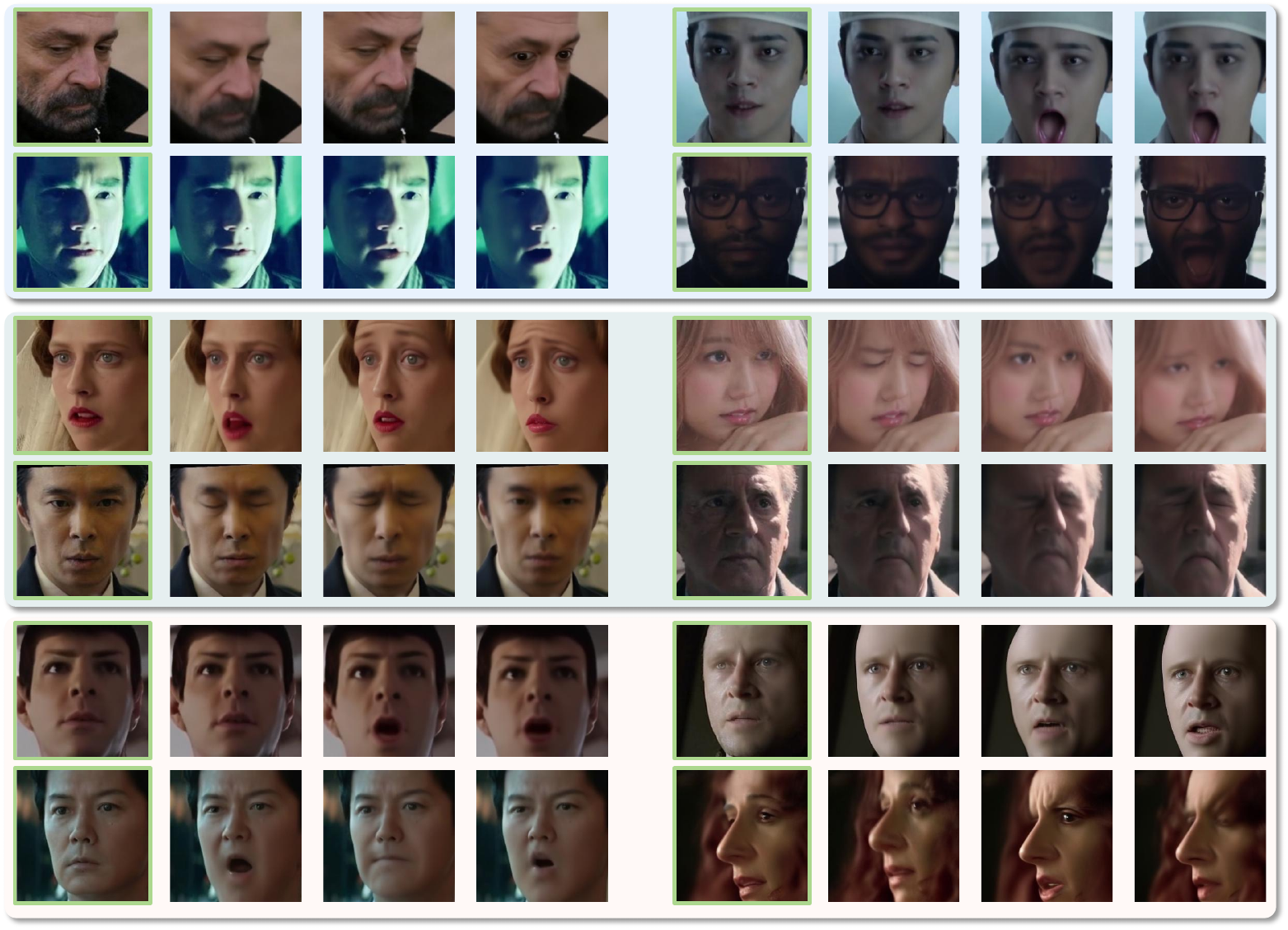}
  \caption{Additional visualizations. The green box indicates the reference frame, while the subsequent three frames represent frames 5, 10, and 15. From top to bottom, represent the emotions of surprise, disgust, and fear.}
  \label{sup2}
\end{figure}

\subsection{Main results}
\textbf{Overall Performance.} 
As shown in Table~\ref{rec1}, ARGen consistently enhances scarce-category recognition across training from scratch, foundation model adaptation, and large-scale pre-training paradigms. It yields substantial gains in UAR and smaller, consistent improvements in WAR. This disparity aligns with the inherent data imbalance: since WAR incorporates sample weighting, it is less sensitive to augmentation given the static test distribution. Notably, ARGen achieves pronounced improvements over the first two data-constrained settings. For the third paradigm, extensive pre-training on millions of external affective videos partially mitigates structural scarcity, narrowing our performance gains.

For generative evaluation, the policy networks optimized on DFEW and FERV39k execute adaptive inference on the laboratory-controlled CK+ dataset. The metrics in Table~\ref{gen1} demonstrate consistent improvements. Critically, sFVD and tFVD deviate from full-scale FVD due to small sample sizes and skewed group distributions, which destabilize covariance estimation and artificially inflate variance and mean values. This statistical artifact does not alter the underlying trajectory, validating our adaptive strategy.

At the dataset level, ARGen impacts DFEW ($\sim$500 samples, $\sim$3.81\% of the training set) more pronouncedly than FERV39k (only 1.6\%). This variance underscores the data-adaptive efficacy of ARGen in counteracting category scarcity without requiring external data inputs.

\begin{table}[t]
\centering
\renewcommand{\arraystretch}{1.2}
\resizebox{0.98\linewidth}{!}{
\begin{tabular}{lccc}
\toprule
Ablation & FVD$\downarrow$ & sFVD$\downarrow$ & tFVD$\downarrow$ \\
\midrule
\textit{ARGen w/o ARD} & & &  \\
\midrule
Fixed prompts (baseline) & 85.87 & 187.55 $\pm$ 75.36 & 112.98 $\pm$ 48.22 \\
VLM-only & 78.40 & 172.10 $\pm$ 63.50 & 108.30 $\pm$ 44.10 \\
\rowcolor{gray!10}
\textbf{VLM+RAG} & \textbf{69.20} & \textbf{157.35 $\pm$ 60.85} & \textbf{92.50 $\pm$ 35.47} \\
\midrule
\textit{ARGen w/o ASI} &  &  &  \\
\midrule
Reinforce-only & 66.80 & 150.18 $\pm$ 55.40 & 90.04 $\pm$ 30.25 \\
Random policy & 90.12 & 205.79 $\pm$ 80.23 & 124.29 $\pm$ 50.42 \\
\rowcolor{gray!10}
\textbf{Learned policy} & \textbf{58.50} & \textbf{136.20 $\pm$ 51.37} & \textbf{87.50 $\pm$ 27.54} \\
\bottomrule
\end{tabular}}
\caption{More detailed ablation experiments on CK+ dataset for ARGen submodules.}
\label{xiaorong}
\end{table}

\textbf{Ablation Studies.} 
As shown in Table~\ref{rec2}, omitting either ASI (employing static prompts) or ARD (excluding the policy network) degrades recognition performance, with ARD yielding the most prominent contribution. Without adaptive strategy constraints, generative quality deteriorates; expanding training data merely mitigates major-class overfitting rather than enhancing scarce-category recognition. 

Table~\ref{gen2} details generative performance under varying reward configurations. Across all settings, a fixed step-count penalty enforces inference efficiency, while facial fidelity and visual quality rewards are evaluated independently. Quality-driven rewards exert a stronger impact, and their joint optimization yields upper-bound performance. Conversely, optimizing solely for facial rewards degrades both inter-frame consistency and global video quality, directly impairing quantitative metrics.

Table~\ref{xiaorong} isolates the granular impact of internal modules. Under the ASI-only configuration, static prompts yield the lowest generative quality due to the absence of affective and identity context. Deploying the VLM alone introduces sparse semantic constraints, resulting in marginal gains, whereas integrating retrieval augmentation (VLM+RAG) provides coherent semantic priors that stabilize outcomes. Under the ARD-only configuration, a random policy exhibits the poorest performance. Optimizing via REINFORCE alone—omitting the guidance curve—achieves moderate improvements based solely on step-count and quality rewards. In contrast, our full policy adaptively optimizes both sampling steps and guidance parameters, significantly mitigating distortion and structural instability.

\textbf{Visualization}.
Figure~\ref{vis0} shows representative generation results, indicating that ARGen produces diverse, high-quality, and efficient samples of scarce expression types while preserving subject identity.
Figure~\ref{sup2} presents further generated results. It can be seen that ARGen's generation results strike a balance between diversity and quality. Additional visualizations are provided in Supplementary Materials Sec.E.


\section{Conclusion}
In this work, we propose ARGen, an affect-reinforced generative augmentation framework to tackle data scarcity and long-tail issues in dynamic emotion perception. ARGen operates via a two-stage pipeline: Affective Semantic Injection (ASI) builds Action Unit (AU)-based knowledge to generate interpretable emotional prompts via vision-language models, while Adaptive Reinforcement Diffusion (ARD) employs multi-dimensional rewards with reinforcement learning to optimize expression naturalness, visual quality, and inference efficiency. Experiments show that ARGen substantially boosts scarce-category recognition accuracy and outperforms diverse baselines, validating the integration of affective priors with adaptive diffusion optimization.

\bibliography{ref.bib}

\end{document}